# Intrinsic3D: High-Quality 3D Reconstruction by Joint Appearance and Geometry Optimization with Spatially-Varying Lighting

Robert Maier[1,2]   Kihwan Kim[1]   Daniel Cremers[2]   Jan Kautz[1]   Matthias Nießner[2,3]
[1]NVIDIA   [2]Technical University of Munich   [3]Stanford University

## Abstract

*We introduce a novel method to obtain high-quality 3D reconstructions from consumer RGB-D sensors. Our core idea is to simultaneously optimize for geometry encoded in a signed distance field (SDF), textures from automatically-selected keyframes, and their camera poses along with material and scene lighting. To this end, we propose a joint surface reconstruction approach that is based on Shape-from-Shading (SfS) techniques and utilizes the estimation of spatially-varying spherical harmonics (SVSH) from subvolumes of the reconstructed scene. Through extensive examples and evaluations, we demonstrate that our method dramatically increases the level of detail in the reconstructed scene geometry and contributes highly to consistent surface texture recovery.*

## 1. Introduction

With the wide availability of commodity RGB-D sensors such as the Microsoft Kinect, Intel RealSense, or Google Tango, reconstruction of 3D scenes has gained significant attention. Along with new hardware, researchers have developed impressive approaches that are able to reconstruct 3D surfaces from the noisy depth measurements of these low-cost devices. A very popular strategy to handle strong noise characteristics is volumetric fusion of independent depth frames [7], which has become the core of many state-of-the-art RGB-D reconstruction frameworks [17, 18, 21, 5, 8].

Volumetric fusion is a fast and efficient solution for regularizing out sensor noise; however, due to its $\ell_2$-regularization property, it tends to oversmooth the reconstruction, leaving little fine-scale surface detail in the result. The same problem also translates to reconstruction of surface textures. Most RGB-D reconstruction frameworks simply map RGB values of associated depth pixels onto the geometry by averaging all colors that have been observed for a given voxel. This typically leads to blurry textures, as wrong surface geometry and misaligned poses introduce re-projection errors where one voxel is associated with dif-ferent color values that are then incorrectly averaged.

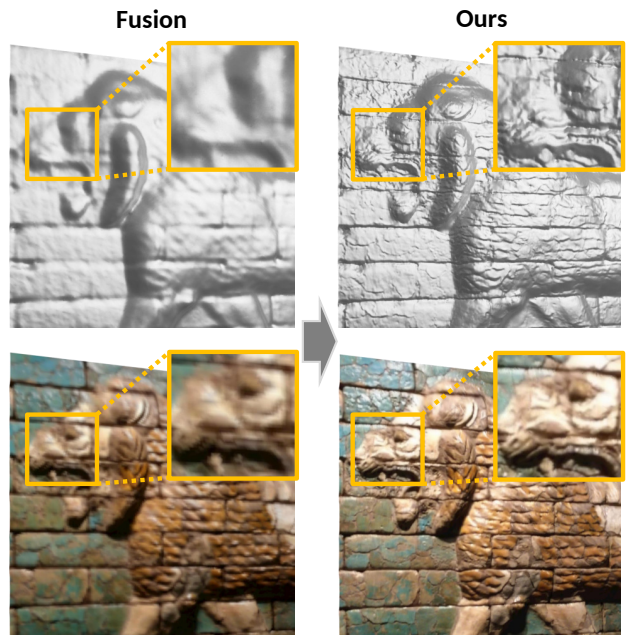

Figure 1. Our 3D reconstruction method jointly optimizes geometry and intrinsic material properties encoded in a Signed Distance Field (SDF), as well as the image formation model to produce high-quality models of fine-detail geometry (top) and compelling visual appearance (bottom).

Very recent approaches address these two problems independently. For instance, Zhou and Koltun [29] optimize for consistent surface textures by iteratively solving for rigid pose alignment and color averages. To compensate for wrong surface geometry where re-projection consistency is infeasible, they non-rigidly warp RGB frames on top of the reconstructed mesh, thus obtaining a high-quality surface texture. On the other end of the spectrum, shading-based refinement techniques enhance depth frames [24] or surface geometry [30] by adding shading constraints from higher resolution color frames; i.e., they leverage RGB signal to refine the geometry. These reconstruction pipelines are sequential; for instance, Zollhöfer et al. [30] first compute the alignment between RGB-D frames, then fuse both RGB and depth data into a volumetric grid, and finally refine the 3D reconstruction. This results in visually promising reconstructions; however, the pipeline fundamentally cannot recover errors in its early stages; e.g., if pose alignment is off due to wrong depth measures, fused colors will be blurry, causing the following geometry refinement to fail.

In our work, we bring these two directions together by addressing these core problems simultaneously rather than separately. Our main idea is to compute accurate surface geometry such that color re-projections of the reconstructed texture are globally consistent. This leads to sharp surface colors, which can again provide constraints for correct 3D geometry. To achieve this goal, we introduce a novel joint optimization formulation that solves for all parameters of a global scene formation model: (1) surface geometry, represented by an implicit signed distance function, is constrained by input depth measures as well as a shading term from the RGB frames; (2) correct poses and intrinsic camera parameters are enforced by global photometric and geometric consistency; (3) surface texture inconsistency is minimized considering all inputs along with the 3D model; and (4) spatially-varying lighting as well as surface albedo values are constrained by RGB measures and surface geometry. The core contribution of our work is to provide a parametric model for all of these intrinsic 3D scene parameters and optimize them in a joint, continuous energy minimization for a given RGB-D sequence. As a result, we achieve both sharp color reconstruction, highly-detailed and physically-correct surface geometry (Figure 1), and an accurate representation of the scene lighting along with the surface albedo. In a series of thorough evaluations, we demonstrate that our method outperforms state-of-the-art approaches by a significant margin, both qualitatively and quantitatively.

To sum up, our technical contributions are as follows:

- We reconstruct a volumetric signed distance function by jointly optimizing for 3D geometry, surface material (albedo), camera poses, camera intrinsics (including lens distortion), as well as accurate scene lighting using spherical harmonics basis functions.
- Instead of estimating only a single, global scene illumination, we estimate spatially-varying spherical harmonics to retrieve accurate scene lighting.
- We utilize temporal view sampling and filtering techniques to mitigate the influence of motion blur, thus efficiently handling data from low-cost consumer-grade RGB-D sensor devices.

## 2. Related Work

**3D Reconstruction using Signed Distance Functions** Implicit surface representations have been widely used in 3D modeling and reconstruction algorithms. In particular, signed distance fields (SDF) [7] are often used to encode 3D surfaces in a voxel grid, and have become the basis of many successful RGB-D surface reconstruction algorithms [17, 18]. More recently, Choi et al. [5] propose a robust optimization for high-quality pose alignment using only geometry, and Dai et al. [8] present a global optimization for large-scale scenes in real time. While most SDF-based fusion methods efficiently regularize noisy depth input, they spend little focus on reconstructing consistent and sharp surface textures. In particular, in the context of wide baseline views and small surface misalignments, this leads to blurry voxel colors that are obtained by averaging the input RGB values of associated color images.

**High-quality texture recovery** In order to compute consistent colors on the reconstructed surface, Zhou and Koltun [29] introduce a method to optimize the mapping of colors onto the geometry (camera poses and 2D deformation grid), Klose et al. [13] propose to filter colors in scene space, and Jeon et al. [12] suggest a more efficient way of color optimization through texture coordinates. In addition to directly optimizing for consistent surface textures, refining texture quality also helps to improve the quality of reconstructed surface colors [16, 9]. While these methods achieve visually impressive RGB reconstructions (e.g., by warping RGB input), they do not address the core problem of color inconsistency, which is caused by wrong surface geometry that leads to inconsistent RGB-to-RGB and RGB-to-geometry re-projections.

**Shading- and reflectance-based geometry refinement** Shape-from-Shading [11, 28] aims to extract 3D geometry from a single RGB image, and forms the mathematical basis of shading-based refinement, targeted by our work. The theory behind Shape-from-Shading is well-studied, in particular when the surface reflectance, light source and camera locations are known. Unfortunately, the underlying optimizations are highly under-constrained, particularly in uncontrolled environments. Thus, one direction is to refine coarse image-based shape models based on incorporation of shading cues [4]. For instance, this can be achieved with images captured by multiple cameras [23, 22] or with RGB-D cameras that provide an initial depth estimate for every pixel [10, 26, 2].

Hence, shading and reflectance estimation has become an important contextual cue for refining geometry. Many methods leverage these cues to develop high-quality surface refinement approaches [24, 19, 3]. In particular, Zollhöfer et al. [30] motivates our direction of using volumetric signed distance fields to represent the 3D model. Unfortunately, the method has significant drawbacks; first, it only assumes a single global lighting setting based on spherical harmonics [20] that is constant over the entire scene; second, its pipeline is sequential, meaning that poses and surface colors are optimized only once in a pre-process, suffering from erroneous depth measures and small pose misalignments. In our approach, we systematically address these shortcomings with a joint optimization strategy, as well as a much more flexible spatially-varying lighting parametrization. Other related methods focus on specular surfaces with an alternating optimization strategy [25], represent lighting with illumination maps [14], or retrieve a box-like 3D representation with material parameters [27].

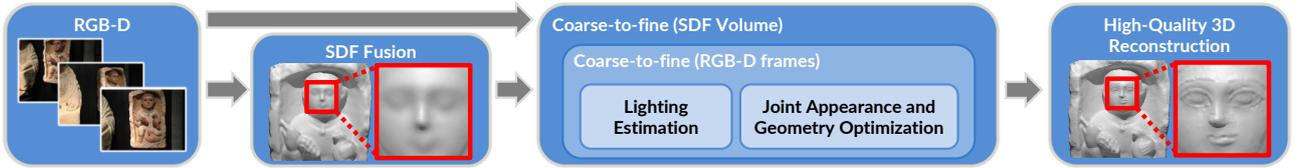

Figure 2. Overview of our method for joint appearance and geometry optimization. Our pipeline takes RGB-D data of a scene as input and fuses it into a Signed Distance Field (SDF). In a nested coarse-to-fine approach, spatially-varying lighting is estimated and used to jointly optimize for appearance and geometry of the scene, producing a high-quality 3D model.

## 3. Overview

Our method first estimates a coarse sparse Signed Distance Field (SDF) similar to Nießner et al. [18] from an input RGB-D sequence with initial camera poses. To mitigate the influence of views with motion blur, we automatically select views based on a blurriness measure and constrain the optimization only based on color values from these keyframes.

Our joint optimization employs a nested hierarchical approach (see Figure 2): in an outer loop, we refine the SDF in a coarse-to-fine manner on multiple SDF grid pyramid levels in order to reconstruct fine detail. At the coarsest grid pyramid level, we use multiple RGB-D frame pyramid levels of all keyframes obtained through downsampling in order to improve the convergence and robustness of the joint camera pose estimation.

Within each inner iteration, we approximate complex scene lighting by partitioning the SDF volume into subvolumes of fixed size with separate spherical harmonics parameters. During estimation, we jointly solve for all SH parameters on a global scale with a Laplacian regularizer. The lighting at a given point is defined as the trilinear interpolation of the associated subvolumes.

In the main stage of our framework, we employ the estimated illumination to jointly refine surface and albedo of the SDF as well as the image formation model (camera poses of the input frames, camera intrinsics and lens distortion). As a consequence of this extensive set of optimized parameters, we implicitly obtain optimal colors. We re-compute the voxel colors from the keyframes using the refined parameters after each optimization. Finally, a 3D mesh is extracted from the refined SDF using Marching Cubes [15].

### 3.1. Signed Distance Field

At the core of our framework lies the reconstructed surface, which we implicitly store as a sparse Truncated Signed Distance Function (TSDF) [7], denoted by $\mathbf{D}$. Hereby, each voxel stores the raw (truncated) signed distance to the closest surface $\mathbf{D}(\boldsymbol{v})$, its color $\mathbf{C}(\boldsymbol{v})$, an integration weight $\mathbf{W}(\boldsymbol{v})$, an illumination albedo $\mathbf{a}(\boldsymbol{v})$, and an optimized signed distance $\tilde{\mathbf{D}}(\boldsymbol{v})$. We denote the current estimate of the iso-surface by $\mathbf{D}_0$ and the number of voxels in the SDF volume by $N$.

Following state-of-the-art reconstruction methods, we integrate depth maps into the SDF using a weighted running average scheme:

$$\mathbf{D}(\boldsymbol{v}) = \frac{\sum_{i=1}^{M} w_i(\boldsymbol{v}) d_i(\boldsymbol{v})}{\mathbf{W}(\boldsymbol{v})}, \quad \mathbf{W}(\boldsymbol{v}) = \sum_{i=1}^{M} w_i(\boldsymbol{v}), \quad (1)$$

with sample integration weight $w_i(\boldsymbol{v}) = \cos(\theta)$, based on the angle $\theta$ between the viewing direction and the normal computed from the input depth map. The truncated signed distance $d_i(\boldsymbol{v})$ between a voxel and a depth frame $\mathcal{Z}_i$ with pose $\mathcal{T}_i$ is computed as follows:

$$d_i(\boldsymbol{v}) = \Psi((\mathcal{T}_i^{-1} \boldsymbol{v})_z - \mathcal{Z}_i(\pi(\mathcal{T}_i^{-1} \boldsymbol{v}))), \quad (2)$$

with truncation $\Psi(d) = \min(|d|, t_{\text{trunc}}) \cdot \text{sgn}(d)$. After integrating all frames of the RGB-D sequence in the implicit 3D model representation, we initialize the optimized SDF $\tilde{\mathbf{D}}$ with the integrated SDF $\mathbf{D}$. We directly compute the surface normal for each voxel from the gradient of the refined signed distance field using forward differences:

$$\mathbf{n}(\boldsymbol{v}) = (n_x, n_y, n_z)^\top = \frac{\nabla \tilde{\mathbf{D}}(\boldsymbol{v})}{||\nabla \tilde{\mathbf{D}}(\boldsymbol{v})||_2}, \quad (3)$$

with the gradient

$$\nabla \tilde{\mathbf{D}}(\boldsymbol{v}) = \nabla \tilde{\mathbf{D}}(i,j,k) = \begin{pmatrix} \tilde{\mathbf{D}}(i+1,j,k) - \tilde{\mathbf{D}}(i,j,k) \\ \tilde{\mathbf{D}}(i,j+1,k) - \tilde{\mathbf{D}}(i,j,k) \\ \tilde{\mathbf{D}}(i,j,k+1) - \tilde{\mathbf{D}}(i,j,k) \end{pmatrix} \quad (4)$$

where $\tilde{\mathbf{D}}(i,j,k)$ is the optimized distance value at the (discrete) voxel location $(i,j,k)$. Since each voxel encodes the distance to its closest surface, it is possible to derive a corresponding 3D point on the iso-surface $\boldsymbol{v}_0$. Thus, the voxel center point $\boldsymbol{v}_c \in \mathbb{R}^3$ in world coordinates is projected onto the (nearest) iso-surface using the transformation $\psi$:

$$\boldsymbol{v}_0 = \psi(\boldsymbol{v}) = \boldsymbol{v}_c - \mathbf{n}(\boldsymbol{v}) \tilde{\mathbf{D}}(\boldsymbol{v}). \quad (5)$$

## 3.2. Image Formation Model and Sampling

**RGB-D Data** As input, our framework takes $M$ RGB-D frames with registered color images $\mathcal{C}_i$, derived intensity images $\mathcal{I}_i$, and depth maps $\mathcal{Z}_i$ (with $i \in 1\dots M$). We assume exposure and white balance of the sensor to be fixed, which is a common setting in RGB-D sensors. Moreover, we are given an initial estimate of the absolute camera poses $\mathcal{T} = \{\mathcal{T}_i\}$ of the respective frames, with $\mathcal{T}_i = (\boldsymbol{R}_i, \boldsymbol{t}_i) \in \mathrm{SE}(3)$, $\boldsymbol{R}_i \in \mathrm{SO}(3)$ and $\boldsymbol{t}_i \in \mathbb{R}^3$. We denote the transformation of a point $\boldsymbol{p}$ using a pose $\mathcal{T}_i$ by $g(\mathcal{T}_i, \boldsymbol{p}) = \boldsymbol{R}_i \boldsymbol{p} + \boldsymbol{t}_i$. While our approach is based on the VoxelHashing framework [18], the initial camera poses can in principle be computed using any state-of-the-art RGB-D based 3D reconstruction system; e.g., [5, 8].

**Camera Model** Our camera model is defined by the focal length $f_x, f_y$, the optical center $c_x, c_y$ and three coefficients $\kappa_1, \kappa_2, \rho_1$ describing radial and tangential lens distortion respectively. 3D points $\boldsymbol{p} = (X, Y, Z)^\top$ are mapped to 2D image pixels $\boldsymbol{x} = (x, y)^\top$ with the projection function $\pi : \mathbb{R}^3 \mapsto \mathbb{R}^2$.

**Keyframe Selection** In hand-held RGB-D scanning, input images often exhibit severe motion blur due to fast camera motion. To mitigate the effect of motion blur, we discard bad views by selecting views using the blurriness measure by Crete et al. [6]. More specifically, we choose the least blurred frame within a fixed size window of $t_{\mathrm{KF}}$ neighboring frames. We set $t_{\mathrm{KF}} = 20$ for regular datasets that are captured with commodity RGB-D sensors, and $t_{\mathrm{KF}} = 5$ for short sequences with less than 100 frames. Our method can also be applied to multi-view stereo datasets consisting of only few images; here, we use all frames (i.e., $t_{\mathrm{KF}} = 1$).

**Observations Sampling and Colorization** After generating the SDF volume, we initially compute the voxel colors by sampling the selected keyframes. Given a frame $(\mathcal{C}_i, \mathcal{Z}_i)$ and its pose $\mathcal{T}_i$, we re-compute the color of a voxel $\boldsymbol{v}$ by sampling its 3D iso-surface point $\boldsymbol{v}_0$ in the input views. To check whether voxel $\boldsymbol{v}$ is visible in view $i$, we transform $\boldsymbol{v}_0$ back into the input view's coordinate system using the (refined) pose $\mathcal{T}_i$, project it into its depth map $\mathcal{Z}_i$ and look up the respective depth value. $\boldsymbol{v}$ is considered visible in the image if the voxel's $z$-coordinate in the camera coordinate system is compatible with the sampled depth value.

We collect all color observations of a voxel in its views and their respective weights in $\mathcal{O}_{\boldsymbol{v}} = \{(c_i^{\boldsymbol{v}}, w_i^{\boldsymbol{v}})\}$. The observed colors $c_i^{\boldsymbol{v}}$ are obtained by sampling from the input color image $\mathcal{C}_i$ using bilinear interpolation:

$$c_i^v = \mathcal{C}_i(\pi(\mathcal{T}_i^{-1} \boldsymbol{v}_0)). \tag{6}$$

The observation weight $w_i^{\boldsymbol{v}}$ is view-dependent on both normal and depth in the view:

$$w_i^{\boldsymbol{v}} = \frac{\cos(\theta)}{d^2}, \tag{7}$$

where $d$ is the distance from $\boldsymbol{v}$ to the camera corresponding to $\mathcal{C}_i$. $\theta$ represents the angle between the voxel normal $\mathbf{n}(\boldsymbol{v})$ rotated into the camera coordinate system, and the view direction of the camera.

**Colorization** We sort the observations in $\mathcal{O}_{\boldsymbol{v}}$ by their weight and keep only the best $t_{\mathrm{best}}$ observations. The voxel color $c_{\boldsymbol{v}}^*$ is computed as the weighted mean of its observations $\mathcal{O}_{\boldsymbol{v}}$ (for each color channel independently):

$$c_{\boldsymbol{v}}^* = \arg\min_{c_{\boldsymbol{v}}} \sum_{(c_i^{\boldsymbol{v}}, w_i^{\boldsymbol{v}}) \in \mathcal{O}_{\boldsymbol{v}}} w_i^{\boldsymbol{v}} (c_{\boldsymbol{v}} - c_i^{\boldsymbol{v}})^2. \tag{8}$$

Note that the per-voxel colors are only used before each optimization step (for up-to-date chromaticity weights) and as a final postprocess during mesh extraction. The optimization itself directly constrains the input RGB images of the selected views and does not use the per-voxel color values.

## 4. Lighting Estimation using Spatially-varying Spherical Harmonics

**Lighting Model** In order to represent the lighting of the scene, we use a fully-parametric model that defines the shading at every surface point w.r.t. global scene lighting. To make the problem tractable, we follow previous methods and assume that the scene environment is Lambertian.

The shading $\mathbf{B}$ at a voxel $\boldsymbol{v}$ is then computed from the voxel surface normal $\mathbf{n}(\boldsymbol{v})$, the voxel albedo $\mathbf{a}(\boldsymbol{v})$ and scene lighting parameters $l_m$:

$$\mathbf{B}(\boldsymbol{v}) = \mathbf{a}(\boldsymbol{v}) \sum_{m=1}^{b^2} l_m H_m(\mathbf{n}(\boldsymbol{v})), \tag{9}$$

with shading basis $H_m$. As Equation 9 defines the forward shading computation, our aim is to tackle the inverse rendering problem by estimating the parameters of $\mathbf{B}$.

**Spherical Harmonics** In order to estimate the reflected irradiance $\mathbf{B}$ (cf. Equation 9) at a voxel $\boldsymbol{v}$, we parametrize the lighting with spherical harmonics (SH) basis functions [20], which is known to be a good approximation and smooth for Lambertian surface reflectance. The SH basis functions $H_m$ are parametrized by a unit normal $\mathbf{n}$. In our implementation, we use SH coefficients up to the second order, which includes $b = 3$ SH bands and leaves us with nine unknown lighting coefficients $\ell = (l_1, \dots, l_{b^2})$. For a given surface point, the SH basis encodes the incident lighting, parameterized as a spherical distribution. However, a single SH basis cannot faithfully represent scene lighting for all surface points simultaneously, as lights are assumed to be infinitesimally far away (i.e., purely directional), and neither visibility nor occlusion is taken into account.

**Subvolume Partitioning** To address the shortcoming of a single, global spherical harmonics basis that globally defines the scene lighting, we extend the traditional formulation. To this end, we partition the reconstruction volume into subvolumes $\mathcal{S} = \{s_1 \ldots, s_K\}$ of fixed size $t_{sv}$; the number of subvolumes is denoted as $K$. We now assign an SH basis – each with its own SH coefficients – to every subvolume. Thus, we substantially increase the number of lighting parameters per scene and allow for spatially-adaptive lighting changes. In order to avoid aliasing artifacts at subvolume boundaries, we define the global lighting function as a trilinear interpolation of local SH coefficients; i.e., for a voxel, we obtain a smooth function defining the actual SH coefficients as an interpolation of the lighting parameters of its eights adjacent subvolumes.

**Spatially-varying Spherical Harmonics** The ability of subvolumes to define local spherical harmonics coefficients along with a global interpolant introduces the concept of spatially-varying spherical harmonics (SVSH). Instead of only representing lighting with a single set of SH coefficients, we have now $K \times b^2$ unknown parameters, that provide for significantly more expressibility in the scene lighting model. The lighting for subvolumes is estimated by minimizing the following objective:

$$E_{\text{lighting}}(\ell_1, \ldots, \ell_K) = E_{\text{appearance}} + \lambda_{\text{diffuse}} E_{\text{diffuse}}. \quad (10)$$

The intuition is that we try to approximate complex global illumination with varying local illumination models for smaller subvolumes. We estimate the spherical harmonics in a subvolume by minimizing the differences between the measured averaged voxel intensity and the estimated appearance:

$$E_{\text{appearance}} = \sum_{\boldsymbol{v} \in \tilde{\mathbf{D}}_0} (\mathbf{B}(\boldsymbol{v}) - \mathbf{I}(\boldsymbol{v}))^2, \quad (11)$$

where only voxels close to the current estimate of the iso-surface $\tilde{\mathbf{D}}_0$ are considered. Initially, we assume the albedo to be constant. However, the albedo is refined as the optimization commences. After the surface refinement on each level, we recompute the voxel colors (and hence voxel intensity). We further regularize the distribution of lighting coefficients with a Laplacian regularizer that considers the 1-ring neighborhood $\mathcal{N}_s$ of a subvolume $s$, thus effectively constraining global smoothness of the spherical harmonics:

$$E_{\text{diffuse}} = \sum_{s \in \mathcal{S}} \sum_{r \in \mathcal{N}_s} (\ell_s - \ell_r)^2. \quad (12)$$

## 5. Joint Optimization of Geometry, Albedo, and Image Formation Model

One of the core ideas of our method is the joint optimization of the volumetric 3D reconstruction as well as the image formation model. In particular, we simultaneously optimize for the signed distance and albedo values of each voxel of the volumetric grid, as well as the camera poses and camera intrinsics such as focal length, center pixel, and (radial and tangential) lens distortion coefficients. We stack all parameters in the unknown vector

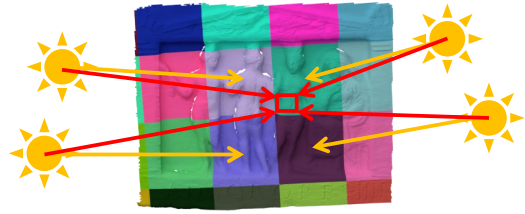

Figure 3. We partition the SDF volume into subvolumes of fixed size and estimate independent spherical harmonics (SH) coefficients for each subvolume (yellow). Per-voxel SH coefficients are obtained through tri-linear interpolation of the lighting of neighboring subvolumes (red).

$\mathcal{X} = (\mathcal{T}, \tilde{\mathbf{D}}, \mathbf{a}, f_x, f_y, c_x, c_y, \kappa_1, \kappa_2, \rho_1)$ and formulate our minimization objective as follows:

$$E_{\text{scene}}(\mathcal{X}) = \sum_{\boldsymbol{v} \in \tilde{\mathbf{D}}_0} \lambda_g E_g + \lambda_v E_v + \lambda_s E_s + \lambda_a E_a, \quad (13)$$

with $\lambda_g, \lambda_v, \lambda_s, \lambda_a$ the weighting parameters that define the influence of each cost term. For efficiency, we only optimize voxels within a thin shell close to the current estimate of the iso-surface $\tilde{\mathbf{D}}_0$, i.e., $|\tilde{\mathbf{D}}| < t_{\text{shell}}$.

### 5.1. Camera Poses and Camera Intrinsics

For initial pose estimates, we use poses obtained by the frame-to-model tracking of VoxelHashing [18]. However, this merely serves as an initialization of the non-convex energy landscape for our global pose optimization, which is performed jointly along with the scene reconstruction (see below). In order to define the underlying residuals of the energy term, we project each voxel into its associated input views by using the current state of the estimated camera parameters. These parameters involve not only the extrinsic poses, but also the pinhole camera settings defined by focal length, pixel center, and lens distortion parameters. During the coarse-to-fine pyramid optimization, we derive the camera intrinsics according to the resolution of the corresponding pyramid levels.

### 5.2. Shading-based SDF Optimization

In order to optimize for the 3D surface that best explains the re-projection and follows the RGB shading cues, we directly solve for the parameters of the refined signed distance field $\tilde{\mathbf{D}}$, which is directly coupled to the shading through its surface normals $\mathbf{n}(\boldsymbol{v})$. In addition to the distance values, the volumetric grid also contains per-voxel albedo parameters, which again is coupled with the lighting computation (cf. Equation 9); the surface albedo is initialized with a uniform constant value. Although this definition of solving for a distance field follows the direction of Zollhöfer et al. [30], it is different at its core: here, we dynamically constrain the reconstruction with the RGB input images, which contrasts Zollhöfer et al. who simply rely on the initially pre-computed per-voxel colors. In the following, we introduce all terms of the shading-based SDF objective.

**Gradient-based Shading Constraint** In our data term, we want to maximize the consistency between the estimated shading of a voxel and its sampled observations in the corresponding intensity images. Our objective follows the intuition that high-frequency changes in the surface geometry result in shading cues in the input RGB images, while more accurate geometry and a more accurate scene formation model result in better sampling of input images.

We first collect all observations in which the iso-surface point $\psi(\boldsymbol{v})$ of a voxel $\boldsymbol{v}$ is visible; we therefore transform the voxel into each frame using the pose $\mathcal{T}_i$ and check whether the sampled depth value in the respective depth map $\mathcal{Z}_i$ is compatible. We collect all valid observations $\mathcal{O}_{\boldsymbol{v}}$, sort them according to their weights $w_i^{\boldsymbol{v}}$ (cf. Equation 7), and keep only the best $t_{\text{best}}$ views $\mathcal{V}_{\text{best}} = \{\mathcal{I}_i\}$. Our objective function is defined as follows:

$$E_g(\boldsymbol{v}) = \sum_{\mathcal{I}_i \in \mathcal{V}_{\text{best}}} w_i^{\boldsymbol{v}} \|\nabla \mathbf{B}(\boldsymbol{v}) - \nabla \mathcal{I}_i(\pi(v_i))\|_2^2, \quad (14)$$

where $v_i = g(\mathcal{T}_i, \psi(\boldsymbol{v}))$ is the 3D position of the voxel center transformed into the view's coordinate system. Observations are weighted with their view-dependent observation weights $w_i^{\boldsymbol{v}}$. By transforming and projecting a voxel $\boldsymbol{v}$ into its associated input intensity images $\mathcal{I}_i$, our joint optimization framework optimizes for all parameters of the scene formation model, including camera poses, camera intrinsics, and lens distortion parameters. The shading $\mathbf{B}(\boldsymbol{v})$ depends on both surface and material parameters and allows to optimize for signed distances, implicitly using the surface normals, and voxel albedo on-the-fly. Instead of comparing shading and intensities directly, we achieve improved robustness by comparing their gradients, which we obtain by discrete forward differences from its neighboring voxels.

To improve convergence, we compute an image pyramid of the input intensity images and run the optimization in a coarse-to-fine manner for all levels. This inner loop is embedded into a coarse-to-fine grid optimization strategy, that increases the resolution of the SDF with each level.

**Regularization** We add multiple cost terms to regularize our energy formulation required for the ill-posed problem of Shape-from-Shading and to mitigate the effect of noise.

First, we use a Laplacian smoothness term to regularize our signed distance field. This volumetric regularizer enforces smoothness in the distance values between neighboring voxels:

$$E_v(\boldsymbol{v}) = (\Delta \tilde{\mathbf{D}}(\boldsymbol{v}))^2. \quad (15)$$

To constrain the surface and keep the refined reconstruction close to the regularized original signed distances, we specify a surface stabilization constraint:

$$E_s(\boldsymbol{v}) = (\tilde{\mathbf{D}}(\boldsymbol{v}) - \mathbf{D}(\boldsymbol{v}))^2. \quad (16)$$

Given spherical harmonics coefficients, the shading computed at a voxel depends on both its albedo as well as its surface normal. We constrain to which degree the albedo or normal should be refined by introducing an additional term that regularizes the albedo. In particular, the 1-ring neighborhood $\mathcal{N}_{\boldsymbol{v}}$ of a voxel is used to constrain albedo changes based on the chromaticity differences of two neighboring voxels. This follows the idea that chromaticity changes often go along with changes of intrinsic material:

$$E_a(\boldsymbol{v}) = \sum_{\boldsymbol{u} \in \mathcal{N}_{\boldsymbol{v}}} \phi(\boldsymbol{\Gamma}(\boldsymbol{v}) - \boldsymbol{\Gamma}(\boldsymbol{u})) \cdot (\mathbf{a}(\boldsymbol{v}) - \mathbf{a}(\boldsymbol{u}))^2, \quad (17)$$

where the voxel chromaticity $\boldsymbol{\Gamma} = \mathbf{C}(\boldsymbol{v})/\mathbf{I}(\boldsymbol{v})$ is directly computed from the voxel colors and $\phi(x)$ is a robust kernel with $\phi(x) = 1/(1 + t_{\text{rob}} \cdot x)^3$.

### 5.3. Joint Optimization Problem

We jointly solve for all unknown scene parameters stacked in the unknown vector $\mathcal{X}$ by minimizing the proposed highly non-linear least squares objective:

$$\mathcal{X}^* = \arg\min_{\mathcal{X}} E_{\text{scene}}(\mathcal{X}) \quad (18)$$

We solve the optimization using the well-known *Ceres Solver* [1], which provides automatic differentiation and an efficient Levenberg-Marquardt implementation.

By jointly refining the SDF and image formation model, we implicitly obtain optimal colors for the reconstruction at minimal re-projection error. In the optimization, the color and shading constraints are directly expressed with respect to associated input images; however, for the final mesh generation, we recompute voxel colors in a postprocess after the optimization. Finally, we extract a mesh from the refined signed distance field using Marching Cubes [15].

## 6. Results

We evaluated our approach on publicly available RGB-D datasets as well as on own datasets acquired using a Structure Sensor; Table 1 gives an overview. For *Lucy* and *Relief* we used the camera poses provided with the datasets as initializations, while we estimated the poses using Voxel Hashing [18] for all other datasets. Our evaluations were performed on a workstation with Intel Core i7-5930 CPU with 3.50GHz and 32GB RAM.

We used $\lambda_{\text{diffuse}} = 0.01, \lambda_g = 0.2, \lambda_v = 160 \to 20, \lambda_s = 120 \to 10, \lambda_a = 0.1$ for our evaluations, with $a \to b$ indicating changing weights with every iteration. For objects with constant albedo, we fixed the albedo; i.e., we set $\lambda_a = \infty$. We used three RGB-D frame pyramid levels and three grid levels, such that the finest grid level has a resolution of 0.5mm (or 1.0mm, depending on object size). We set $t_{\text{best}} = 5$ to limit the number of data term residuals per voxel. To reduce the increase of the number of voxels close to the surface considered for optimization, we used an adaptive thin shell size $t_{\text{shell}}$, linearly decreasing

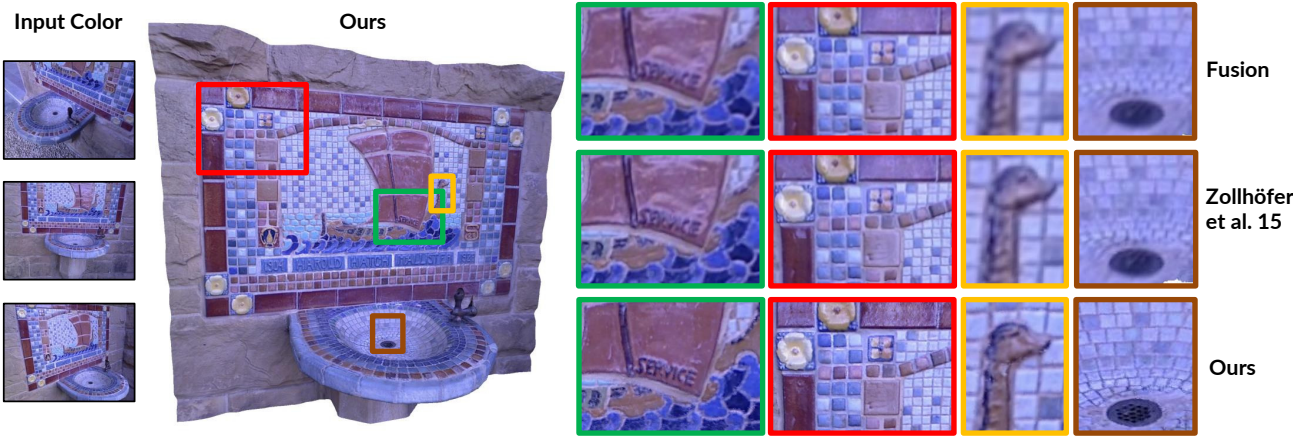

Figure 4. Appearance of the *Fountain* reconstruction. Our method shows a visually more appealing result compared to volumetric fusion and Zollhöfer et al. [30].

| Dataset | # frames | # keyframes | Resolution | |
| --- | --- | --- | --- | --- |
| | | | color | depth |
| *Fountain* [29] | 1086 | 55 | 1280x1024 | 640x480 |
| *Lucy* [30] | 100 | 20 | 640x480 | 640x480 |
| *Relief* [30] | 40 | 8 | 1280x1024 | 640x480 |
| *Lion* | 515 | 26 | 1296x968 | 640x480 |
| *Tomb Statuary* | 523 | 27 | 1296x968 | 640x480 |
| *Bricks* | 773 | 39 | 1296x968 | 640x480 |
| *Hieroglyphics* | 919 | 46 | 1296x968 | 640x480 |
| *Gate* | 1213 | 61 | 1296x968 | 640x480 |

Table 1. Test RGB-D datasets used for the evaluation.

from $2.0 \rightarrow 1.0$ times the voxel size with each grid pyramid level.

**Appearance**  Using our method, we implicitly obtain optimal voxel colors as a consequence of the joint optimization of intrinsic material properties, surface geometry and image formation model. Figure 4 shows qualitative results from the *Fountain* dataset. While volumetric blending [17, 18] produces blurry colors, camera poses are corrected in advance by Zollhöfer et al. [30] using dense bundle adjustment to yield significantly better color and geometry. However, their static color integration cannot correct for small inaccuracies, resulting in slightly blurry colors. In contrast, our method adjusts the surface and image formation model jointly to produce highly detailed texture at the same voxel grid resolution of 1mm. Within our joint optimization, we also estimate varying albedo. Figure 7 shows the estimated albedo for the *Fountain* dataset.

**Surface Geometry**  We qualitatively compare the quality of refined surfaces using our method with the approach of Zollhöfer et al. [30] in Figure 5. The results of the *Relief* dataset visualize that our method reveals finer geometric details by directly sampling from high-resolution input color images instead of using averaged voxel colors. Moreover, we benefit from simultaneously optimizing for camera poses and camera intrinsics.

Additionally, we provide a quantitative ground truth evaluation of the geometry refinement on the synthetic *Frog*

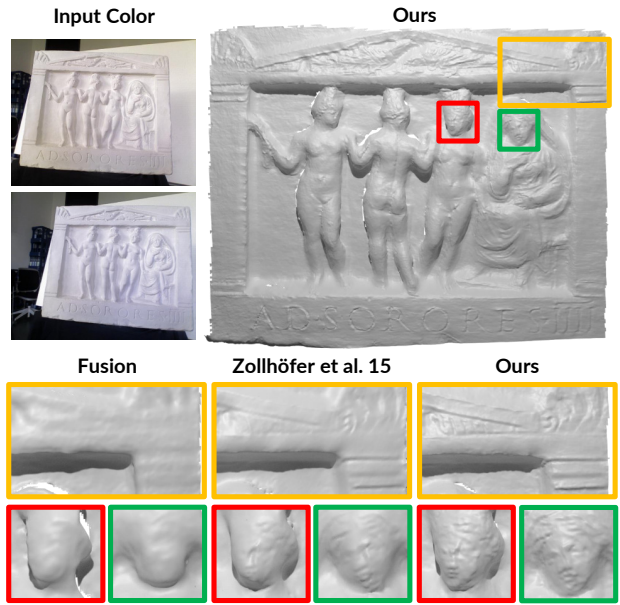

Figure 5. Comparison of the reconstructed geometry of the *Relief* dataset. Our method (right) reveals finer geometric details compared to volumetric fusion (left) and Zollhöfer et al. [30] (middle).

RGB-D dataset, which was generated by rendering a ground truth mesh with a high level of detail into synthetic color and depth images. Both depth and camera poses were perturbed with realistic noise. Figure 6 shows that, in contrast to fusion and [30], our method is able to reveal even smaller details. Quantitatively, the mean absolute deviation (MAD) between our reconstruction and the ground truth mesh is 0.222mm (with a standard deviation of 0.269mm), while the reconstruction generated using our implementation of [30] results in a higher error of 0.278mm (with a standard deviation of 0.299mm). This corresponds to an overall accuracy improvement of $20.14\%$ of our method compared to [30]. We refer the reader to the appendix for a quantitative evaluation on real data and further results.

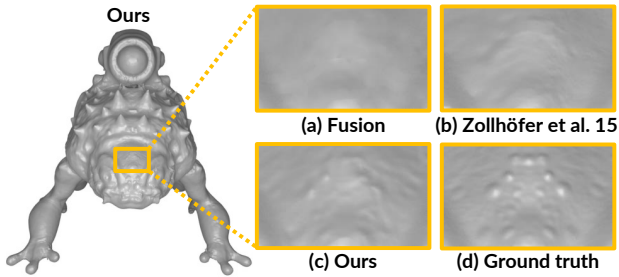

Figure 6. Refined geometry of the *Frog* dataset: while fusion (a) smooths out high-frequency details, Zollhöfer et al. [30] (b) can reconstruct some geometric details. Our method (c) recovers even smaller surface details present in the ground truth mesh (d).

| Dataset | Global SH | SVSH (subvolume size) | | | |
|---|---|---|---|---|---|
| | | 0.5 | 0.2 | 0.1 | 0.05 |
| *Fountain* | 22.973 | 18.831 | 15.891 | 13.193 | **10.263** |
| *Lucy* | 22.190 | 19.408 | 16.564 | 14.141 | **11.863** |
| *Relief* | 13.818 | 12.432 | 11.121 | 9.454 | **8.339** |
| *Lion* | 30.895 | 25.775 | 20.811 | 16.243 | **13.468** |
| *Tomb Statuary* | 33.716 | 30.873 | 30.639 | 29.675 | **26.433** |
| *Bricks* | 29.327 | 27.110 | 25.318 | 22.850 | **19.476** |
| *Hieroglyphics* | 15.710 | 15.206 | 11.140 | 12.448 | **9.998** |
| *Gate* | 46.463 | 40.104 | 33.045 | 20.176 | **12.947** |

Table 2. Quantitative evaluation of spatially-varying spherical harmonics. The Mean Absolute Deviation (MAD) between averaged per-voxel intensity and estimated shading decreases with decreasing subvolume sizes.

**Lighting** In the following, we evaluate lighting estimation via spatially-varying spherical harmonics, both qualitatively and quantitatively. In particular, we demonstrate that a single global set of SH coefficients cannot accurately reflect real-world environments with complex lighting. To analyze the effects of the illumination, we re-light the reconstruction using the surface normals and estimated voxel albedo according to Equation 9. The computed shading $\mathbf{B}(\bm{v})$ of a voxel is in the ideal case identical to the measured voxel intensity $\mathbf{I}(\bm{v})$ computed from the voxel color.

We exploit the absolute difference $|\mathbf{B}(\bm{v}) - \mathbf{I}(\bm{v})|$ as an error metric in order to quantitatively evaluate the quality of the illumination for given geometry and albedo. In particular, we measure the mean absolute deviation (MAD) for all $N$ voxels of the SDF volume:

$$\epsilon_{\text{shading}} = \frac{1}{N} \sum_{\bm{v} \in \mathbf{D}} |\mathbf{B}(\bm{v}) - \mathbf{I}(\bm{v})| \qquad (19)$$

Table 2 gives the results of global SH coeffecints and SVSH with varying subvolume sizes for multiple datasets. In summary, the more the SDF volume is partitioned into subvolumes, the better the approximation to complex lighting scenarios. The illumination in the *Fountain* dataset is clearly spatially varying, violating the assumptions of distant and spatially invariant illumination for SH lighting coefficients. Figure 7 shows that the estimated shading is better approximated with SVSH coefficients compared to only with global SH coefficients, while the underlying surface and albedo are exactly the same for both shadings.

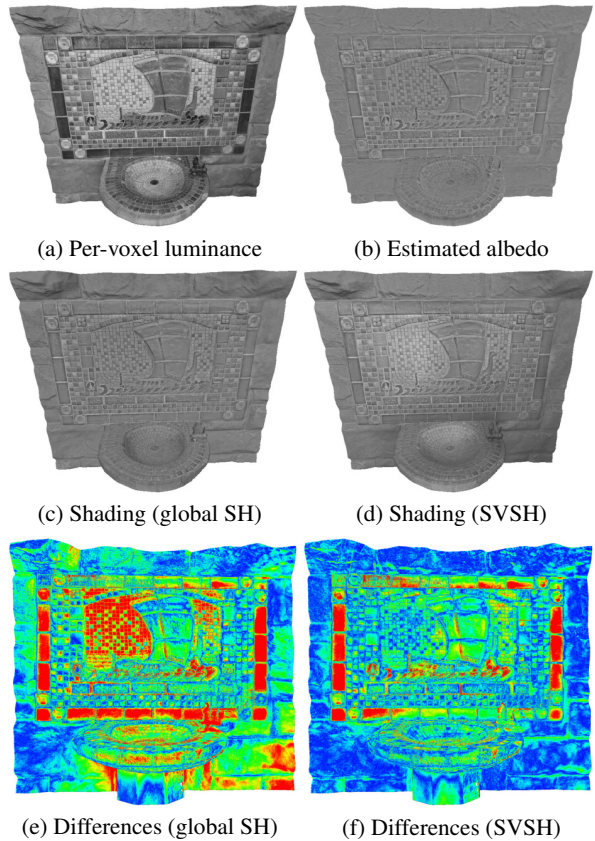

(a) Per-voxel luminance  (b) Estimated albedo

(c) Shading (global SH)  (d) Shading (SVSH)

(e) Differences (global SH)  (f) Differences (SVSH)

Figure 7. Quantitative evaluation of global SH vs. SVSH: the heatmaps in (e) and (f) represent the differences between the per-voxel input luminance (a) and the shadings with global SH (c) and with SVSH (d), both with underlying albedo (b).

## 7. Conclusion

We have presented a novel method for simultaneous optimization of scene reconstruction along with the image formation model. This way, we obtain high-quality reconstructions along with well-aligned sharp surface textures using commodity RGB-D sensors by efficiently combining information from (potentially noisy) depth and (possibly) higher resolution RGB data. In comparison to existing Shape-from-Shading techniques (e.g., [24, 30]), we tackle the core problem of fixing wrong depth measurements jointly with pose alignment and intrinsic scene parameters. Hence, we minimize re-projection errors, thus avoiding oversmoothed geometry and blurry surface textures. In addition, we introduce a significantly more flexible lighting model that is spatially-adaptive, thus allowing for a more precise estimation of the scene lighting.

**Acknowledgment** We would like to thank Qian-Yi Zhou and Vladlen Koltun for the *Fountain* data and Michael Zollhöfer for the *Socrates* laser scan. This work was partially funded by the ERC Consolidator grant *3D Reloaded*.

# Appendix

In this appendix, we provide additional experiments and details. Specifically, we give an overview of the mathematical symbols in Sec. A, and in Sec. B we provide a thorough quantitative evaluation regarding the geometric reconstruction quality on ground truth data (both real and synthetic). We further show qualitative results of the reconstructed models on several own and publicly-available datasets, with a focus on both reconstruction geometry and appearance; see Sec. C. Finally, in Sec. D, we detail additional experiments on spatially-varying lighting under both qualitative and quantitative standpoints.

## A. List of Mathematical Symbols

| Symbol | Description |
| --- | --- |
| $\boldsymbol{p}$ | continuous 3D point in $\mathbb{R}^3$ |
| $\boldsymbol{x}$ | continuous 2D image point in $\mathbb{R}^2$ |
| $\boldsymbol{v}$ | position of voxel in $\mathbb{R}^3$ |
| $\boldsymbol{v}_c$ | position of voxel center of $\boldsymbol{v}$ in $\mathbb{R}^3$ |
| $\boldsymbol{v}_0$ | position of $\boldsymbol{v}$ transformed onto iso-surface in $\mathbb{R}^3$ |
| $\mathbf{n}(\boldsymbol{v})$ | surface normal at $\boldsymbol{v}$ in $\mathbb{R}^3$ |
| $\mathbf{D}(\boldsymbol{v})$ | signed distance value at $\boldsymbol{v}$ |
| $\mathbf{C}(\boldsymbol{v}), \mathbf{I}(\boldsymbol{v})$ | color (RGB) and intensity at $\boldsymbol{v}$ |
| $\mathbf{W}(\boldsymbol{v})$ | integration weight at $\boldsymbol{v}$ |
| $\mathbf{a}(\boldsymbol{v})$ | albedo at $\boldsymbol{v}$ |
| $\tilde{\mathbf{D}}(\boldsymbol{v})$ | refined signed distance value at $\boldsymbol{v}$ |
| $\mathbf{D}_0$ | iso-surface of the refined SDF |
| $\mathbf{B}(\boldsymbol{v})$ | estimated reflected shading at $\boldsymbol{v}$ |
| $\boldsymbol{\Gamma}(\boldsymbol{v})$ | chromaticity at $\boldsymbol{v}$ |
| $t_{\text{shell}}$ | thin shell size |
| $N$ | number of voxels inside the thin shell region |
| $K, t_{\text{sv}}$ | number of subvolumes and subvolume size in $\mathbb{R}^3$ |
| $\mathcal{S}$ | set of subvolumes $s_k$ |
| $\ell$ | vector of all lighting coefficients $l_m$ |
| $H_m$ | $m$-th spherical harmonics basis |
| $b$ | number of spherical harmonics bands |
| $M$ | number of input frames |
| $\mathcal{C}_i, \mathcal{I}_i, \mathcal{Z}_i$ | color, intensity and depth image of frame $i$ |
| $\mathcal{T}_i$ | transformation from frame $i$ to the base frame |
| $t_{\text{KF}}$ | keyframe selection window size |
| $t_{\text{best}}, \mathcal{V}_{\text{best}}$ | number of best views for $\boldsymbol{v}$ and corresponding set |
| $d_i(\boldsymbol{v})$ | projective distance to voxel center in frame $i$ |
| $w_i(\boldsymbol{v})$ | sample integration weight of frame $i$ |
| $\mathcal{O}_{\boldsymbol{v}}$ | set of color observations of $\boldsymbol{v}$ |
| $c_i^{\boldsymbol{v}}$ | observed color of $\boldsymbol{v}$ in frame $i$ |
| $w_i^{\boldsymbol{v}}$ | observation weight of $\boldsymbol{v}$ in frame $i$ |
| $f_x, f_y, c_x, c_y$ | camera intrinsics (focal length, optical center) |
| $\kappa_1, \kappa_2, \rho_1$ | radial and tangential lens distortion parameters |
| $\mathcal{X}$ | stacked vector of optimization variables |

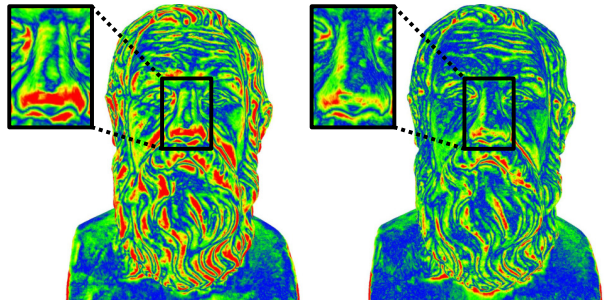

Figure 8. Surface accuracy comparison with a ground truth laser scan of the *Socrates* dataset: the approach of Zollhöfer et al. [30] (left) exhibits a higher mean absolute deviation from the ground truth compared to our method (right).

## B. Quantitative Geometry Evaluation

In the following, we show a quantitative surface accuracy evaluation of our geometry refinement on the *Socrates* and *Frog* datasets.

### B.1. Socrates

In order to measure the surface accuracy of our method quantitatively, we first compare our method with a ground truth laser scan of the *Socrates* Multi-View Stereo dataset from [30]. The mean absolute deviation (MAD) between our reconstruction and the laser scan is 1.09mm (with a standard deviation of 2.55mm), while the publicly-available refined 3D model of Zollhöfer et al. [30] has a significantly higher mean absolute deviation of 1.80mm (with a standard deviation of 3.35mm). This corresponds to an accuracy improvement of $39.44\%$ of our method. Figure 8 visualizes the color-coded mean absolute deviation on the surface.

### B.2. Frog

Besides a quantitative comparison with a laser scan, we also evaluate the surface accuracy of a 3D model reconstructed from synthetic RGB-D data. We therefore generated the synthetic *Frog* dataset by rendering a ground truth mesh with a high level of detail into synthetic color and depth images. We smooth the depth maps using a bilateral filter and add Gaussian noise to both the depth values and to the camera poses.

Instead of comparing the reconstructed 3D models directly with the original mesh, we instead fuse the generated noise-free RGB-D frames into a Signed Distance Field and extract a 3D mesh with Marching Cubes [15]. This extracted mesh is then used as ground truth reference and represents the best possible reconstruction given the raycasted input data in combination with an SDF volume representation.

The mean absolute deviation between our reconstruction and the ground truth mesh is 0.222mm (with a standard de-viation of 0.269mm). With the reconstruction generated using our implementation of [30], we obtain a substantially higher mean absolute deviation of 0.278mm (with a standard deviation of 0.299mm). Compared to [30], our method improves the reconstruction accuracy by 20.14% and is able to reveal geometric details lost with [30]. Figure 9 visualizes the color-coded mean absolute deviation on the surface.

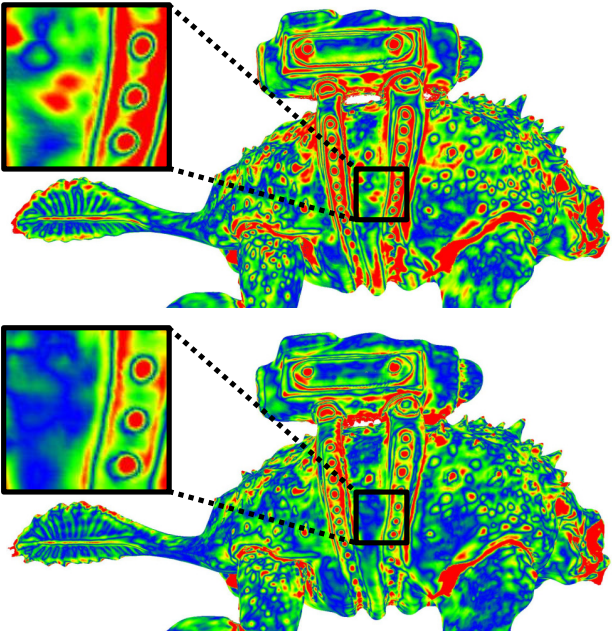

Figure 9. Surface accuracy comparison on synthetic data with a ground truth mesh of the *Frog* dataset: our method (bottom) generates more accurate results compared to Zollhöfer et al. [30] (top).

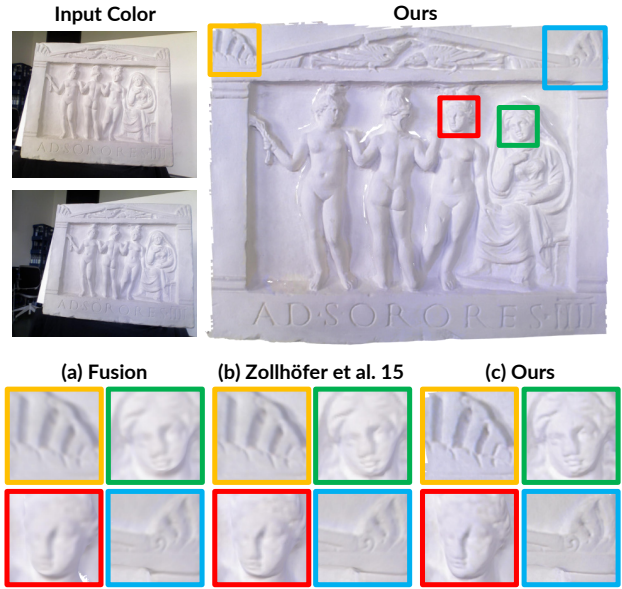

Figure 10. Refined appearance of *Relief* dataset: our method (c) reconstructs significantly sharper textures compared to (a) and (b). Close-ups of ornaments (yellow, blue) and figures (green, red) exhibit more visual details.

## C. Examples of 3D Reconstructions

In addition to providing a thorough quantitative ground truth evaluation, we show qualitative results of 3D models reconstructed from several RGB-D datasets. In particular, we present 3D reconstructions of the publicly-available *Relief* and *Lucy* datasets from Zollhöfer et al. [30] as well as 3D models of the *Gate*, *Lion*, *Hieroglyphics*, *Tomb Statuary* and *Bricks* datasets that we acquired with a Structure Sensor.

Apart from showing the fine detailed geometry, we also demonstrate the improved appearance of the reconstructions, which we implicitly obtain by jointly optimizing for surface, albedo, and image formation model parameters within our approach.

### C.1. Relief

In Figure 10, we show a comparison of the appearance generated using our method with simple volumetric fusion (e.g., Voxel Hashing [18]) and the shading-based surface refinement approach by Zollhöfer et al. [30]. The results in (a) and (b) are visualizations from the meshes that are publicly-available on the project website of [30]. The close-ups successfully visualize that our method results in significantly sharper textures.

### C.2. Lucy

In Figure 11, we present a visual comparison of the reconstructed surface geometry of the *Lucy* dataset. Note how volumetric fusion (a) and Zollhöfer et al. [30] (b) cannot reveal fine-scale details due to the use of averaged per-voxel colors for the refinement, while our method gives the best results and provides geometric consistency (c).

Regarding appearance, we can observe in Figure 12 that our method (c) provides a more detailed texture compared to fusion (a) and Zollhöfer et al. [30] (b).

### C.3. Additional Datasets

While the *Relief* and *Lucy* datasets provided by [30] consist of rather small objects with only few input RGB-D frames and short camera trajectories, we acquired more advanced RGB-D datasets using a Structure Sensor.

Figure 13 shows the reconstruction of the *Gate* dataset, while the 3D model of the *Lion* dataset is visualized in Figure 14. The 3D reconstructions of *Hieroglyphics*, *Tomb Statuary* and *Bricks* are presented in Figure 15, Figure 16 and Figure 17 respectively. For all of these datasets, our method generates high-quality 3D reconstructions with fine-scale surface details and and compelling visual appearance with sharp texture details. In contrast, the models ob-tained from volumetric fusion lack fine details in both geometry and appearance.

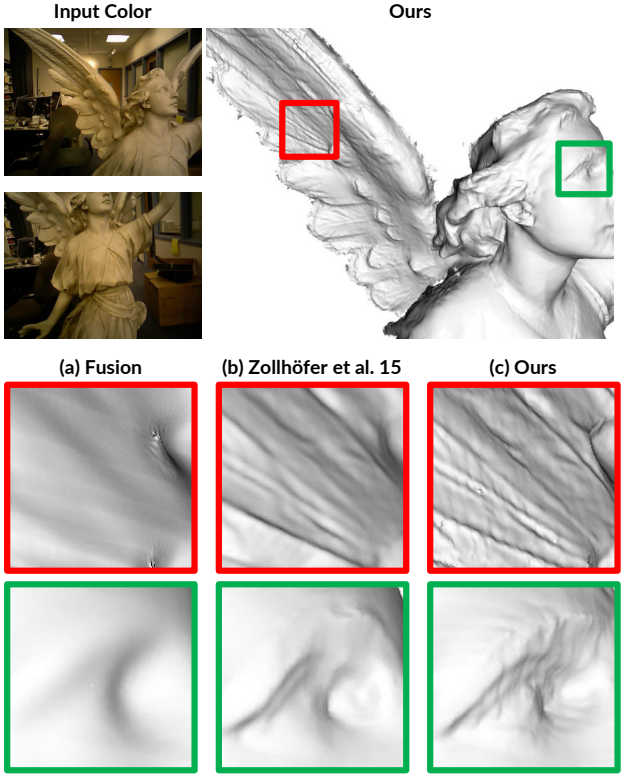
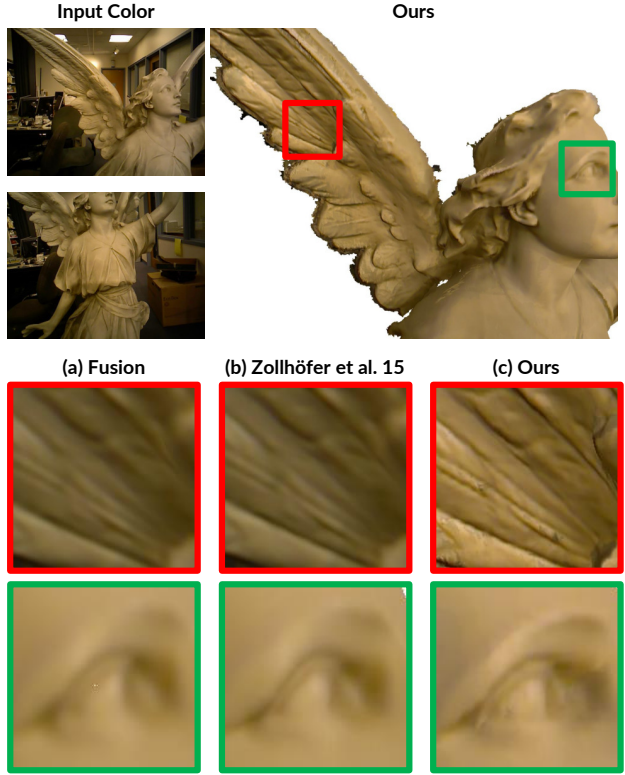

Figure 11. Refined geometry of *Lucy* dataset: volumetric fusion (a) with its strong regularization gives only coarse models. Zollhöfer et al. [30] (b) generate more details; however, limited by using averaged per-voxel colors for the refinement. Our approach that jointly optimizes for all involved parameters (c) reconstructs fine-detailed high-quality geometry.

Figure 12. Refined appearance of *Lucy* dataset: in addition to precise geometry our method (c) also produces high-quality colors compared to (a) and (b).

## D. Evaluation of Spatially-Varying Lighting

In this section, we present further qualitative results for lighting estimation via spatially-varying spherical harmonics (SVSH) compared to global spherical harmonics (global SH) on various datasets. We use the same underlying geometry for both variants of lighting estimation for each dataset.

**Error Metric** As a metric, we use the absolute difference between estimated shading and observed input luminance of a voxel $v$; i.e.,

$$\mathbf{B}_{\text{diff}} = |\mathbf{B}(v) - \mathbf{I}(v)|, \tag{20}$$

to determine the quality of the illumination for given geometry and albedo. Ideally, this difference should be as small as possible.

**Relief** For the *Relief* dataset, the differences between lighting estimation with global SH and SVSH (with a subvolume size of 0.05m) are shown in Figure 18. It becomes obvious that even for seemingly simple scenes, a single global set of Spherical Harmonics coefficients cannot accurately reflect real-world environments with complex lighting.

**Lucy** Similar to the *Relief*, SVSH (with a subvolume size of 0.05m) can better approximate the complex illumination in the *Lucy* dataset than global SH. Figure 19 visualizes the differences in the estimated shadings.

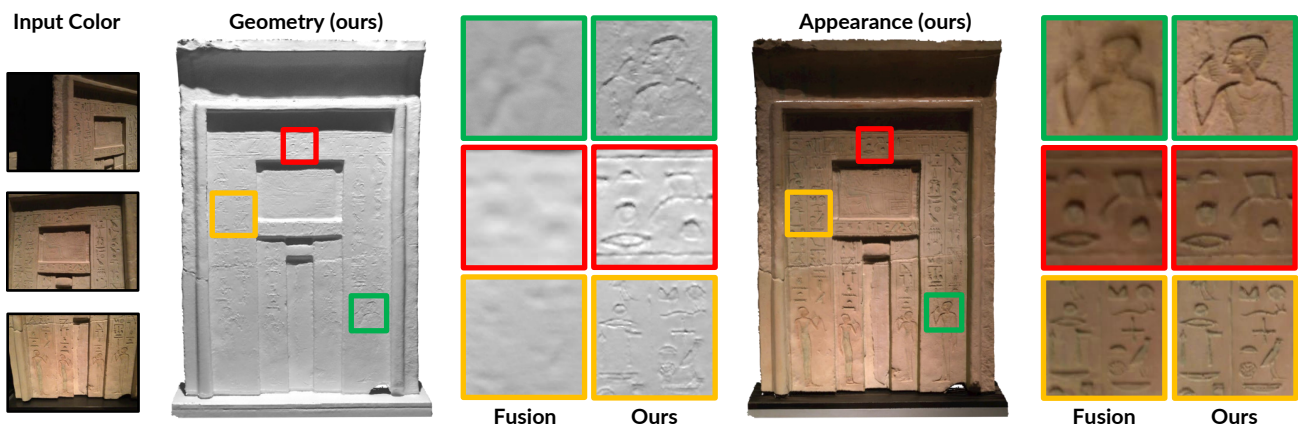

Figure 13. Reconstruction of the *Gate* dataset.

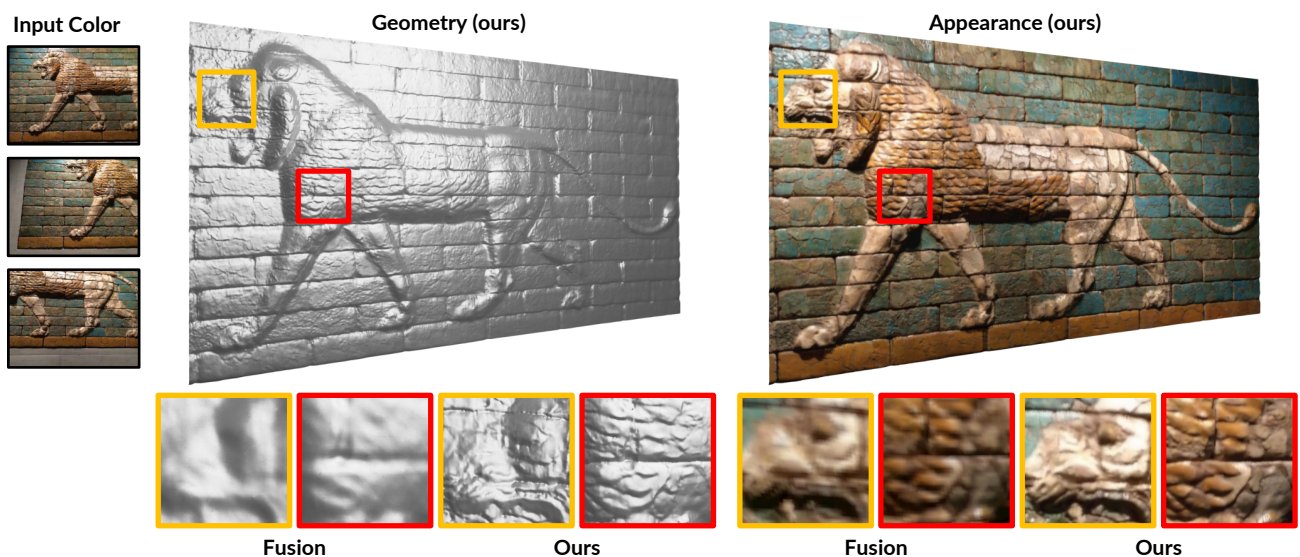

Figure 14. Reconstruction of the *Lion* dataset.

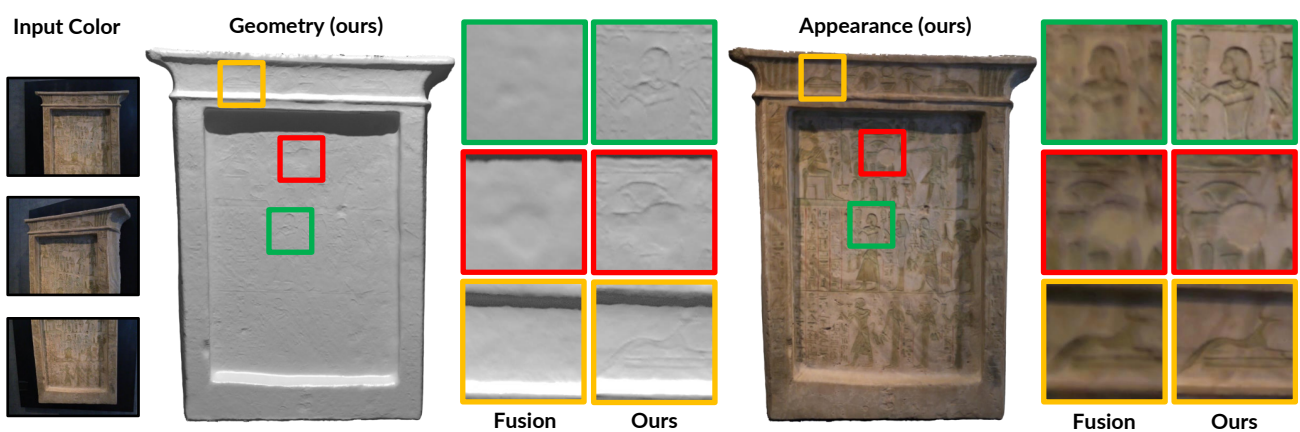

Figure 15. Reconstruction of the *Hieroglyphics* dataset.

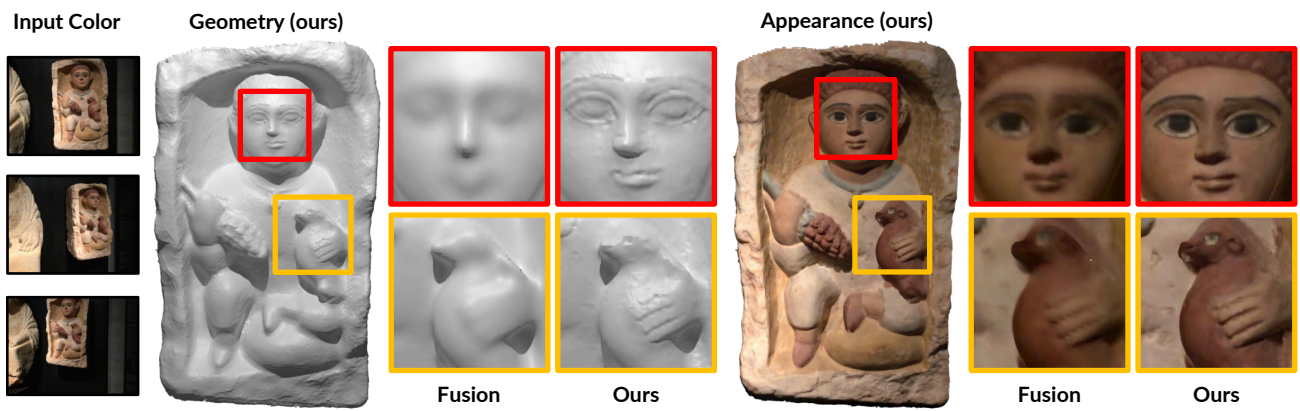

Figure 16. Reconstruction of the *Tomb Statuary* dataset.

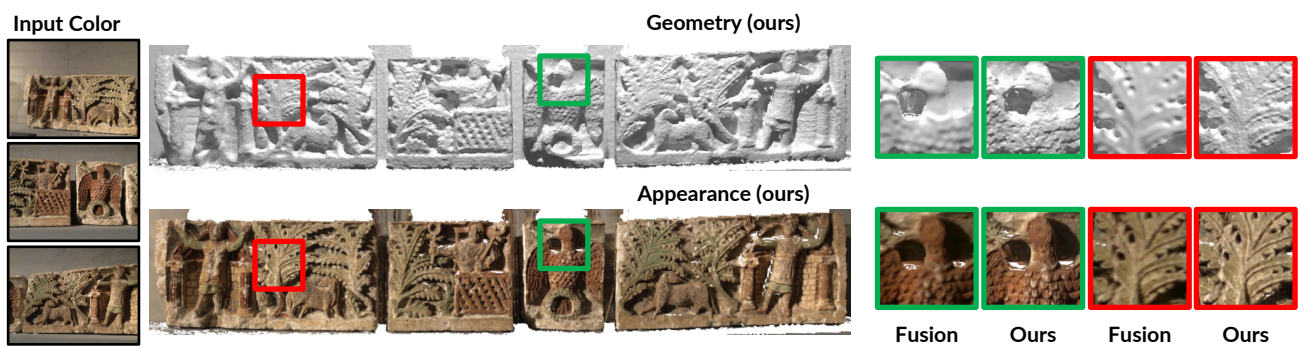

Figure 17. Reconstruction of the *Bricks* dataset.

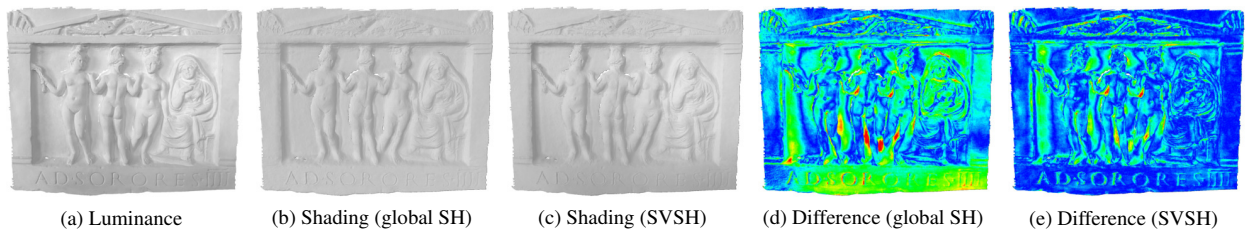

(a) Luminance  (b) Shading (global SH)  (c) Shading (SVSH)  (d) Difference (global SH)  (e) Difference (SVSH)

Figure 18. Estimated illumination of *Relief* dataset: the differences between input luminance (a) and estimated shading (b) and (c) are less for SVSH (e) than for global SH (d), meaning a better approximation of the illumination.

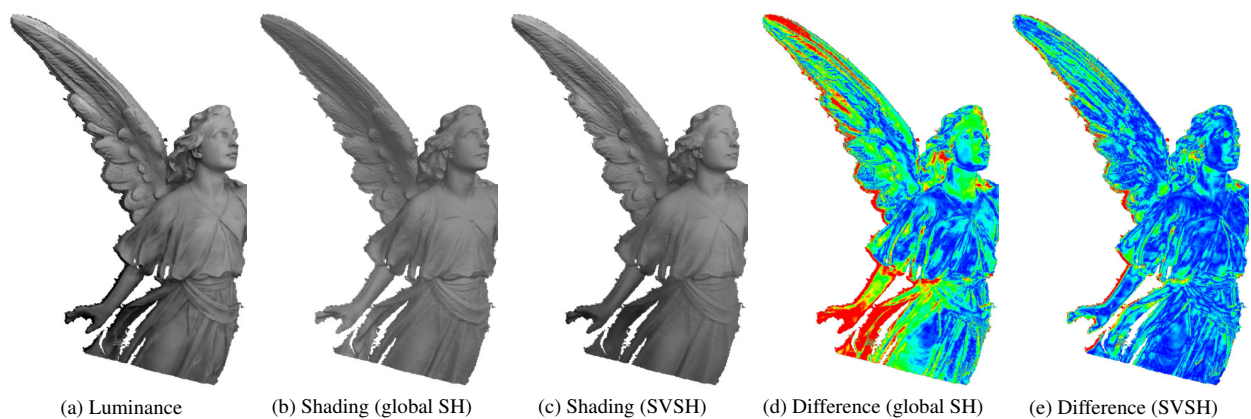

(a) Luminance  (b) Shading (global SH)  (c) Shading (SVSH)  (d) Difference (global SH)  (e) Difference (SVSH)

Figure 19. Estimated illumination of *Lucy* dataset: illumination with SVSH (c) explains the illumination better than global SH only (b), resulting in less differences (e) compared to (d) between input luminance (a) and shading.

14